\documentclass{ieeeaccess}
\usepackage{amsmath,amssymb,amsfonts}
\usepackage{algorithmic}
\usepackage{graphicx}
\usepackage[numbers]{natbib}
\usepackage{textcomp}
\makeatletter
\usepackage{framed,multirow}
\usepackage{amssymb}
\usepackage{latexsym}
\usepackage{url}
\usepackage{caption}
\usepackage{subfig}
\usepackage{times}
\usepackage{epsfig}
\usepackage{graphicx}
\usepackage{amsmath}
\usepackage{amssymb}
\usepackage{bold-extra}

\usepackage{makecell}
\usepackage[x11names,dvipsnames,table]{xcolor}
\usepackage{framed,multirow}
\usepackage{atbegshi}
\newcounter{magicrownumbers}
\newcommand\rownumber{\stepcounter{magicrownumbers}\arabic{magicrownumbers}}
\newcommand{\argmax}{\operatornamewithlimits{argmax}}
\usepackage[pagebackref=false,breaklinks=false,letterpaper=true,colorlinks,bookmarks=false]{hyperref}
\usepackage{wrapfig}
\def\BibTeX{{\rm B\kern-.05em{\sc i\kern-.025em b}\kern-.08em
 T\kern-.1667em\lower.7ex\hbox{E}\kern-.125emX}}
\begin{document}
\history{Received 26 May 2022, accepted 21 June 2022, date of publication 27 June 2022, date of current version 13 July 2022.}
\doi{10.1109/ACCESS.2022.3186471}

\title{EKTVQA: Generalized use of External Knowledge to empower Scene Text in \mbox{Text-VQA}}
\author{\uppercase{Arka Ujjal Dey}\authorrefmark{1}, 
\uppercase{Ernest Valveny \authorrefmark{2}, and Gaurav Harit}\authorrefmark{1}}
\address[1]{IIT Jodhpur, Rajasthan, India}
\address[2]{Computer Vision Center, Universitat Aut\`{o}noma de Barcelona, Bellaterra (Barcelona)}
\tfootnote{This work was supported by Scheme for Promotion of Academic and Research Collaboration (SPARC), Ministry of Human Resource Development (MHRD), Government of India.}

\markboth
{A. Dey \headeretal: EKTVQA: Generalized use of External Knowledge to empower Scene Text in \mbox{Text-VQA}}
{A. Dey \headeretal: EKTVQA: Generalized use of External Knowledge to empower Scene Text in \mbox{Text-VQA}}

\corresp{Corresponding author: Arka Ujjal Dey (e-mail: dey.1@iitj.ac.in).}

\begin{abstract}
The open-ended question answering task of \mbox{Text-VQA} often requires reading and reasoning about \emph{rarely seen or completely unseen} scene text content of an image. We address this zero-shot nature of the task by proposing the generalized use of external knowledge to augment our understanding of the scene text. 
We design a framework to extract, validate, and reason with knowledge using a standard multimodal transformer for vision language understanding tasks. Through empirical evidence and qualitative results, we demonstrate how external knowledge can highlight instance-only cues and thus help deal with training data bias, improve answer entity type correctness, and detect multiword named entities. We generate results comparable to the state-of-the-art on three publicly available datasets under the constraints of similar upstream OCR systems and training data. 
\end{abstract}


\titlepgskip=-15pt

\maketitle
\section{Introduction}\label{sec:intro}
\begin{figure}
\begin{center}
\includegraphics[width=0.45\textwidth]{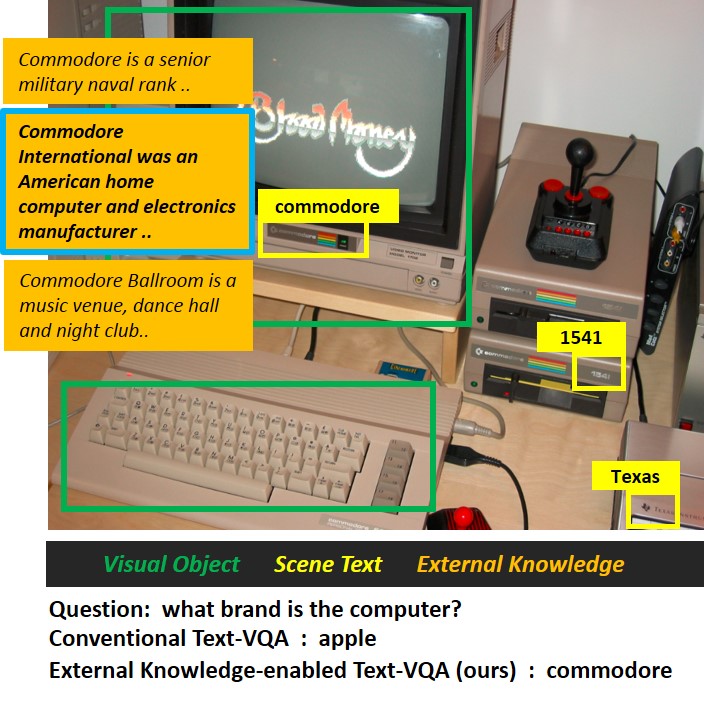}
\caption{Core idea -- Use context validated external knowledge fact, marked blue, along with detected visual and scene text objects to answer questions about an image.}\label{fig:vsem_main}
\end{center}
\end{figure}
Scene text is ubiquitous in our everyday images~\cite{dey2019visual}, from street or store names~\cite{me_karaoglu2017words} to advertisements~\cite{ad_hussain2017automatic} or brand logos. However, early works in visual understanding tasks like Visual Question Answering (VQA)~\cite{antol2015vqa} and Captioning \cite{BAI2018291} did not leverage the scene text. Scene text recognition was still in its early stages, and manual annotation of text is resource-intensive. Recent advances in scene text recognition~\cite{chen2021text} have improved their ability to read the text in natural images, making the scene text content more accessible.
This has opened up new language vision tasks and datasets, like \mbox{Text-VQA}~\cite{tmg_singh2019textvqa,tmg_biten2019scene} and Text-Caption~\cite{sidorov2020textcaps}, that explicitly requires us to read and reason with the scene text of an image. However, the scene text words come from a long-tailed distribution, giving such tasks \emph{zero-shot} characteristics. We hypothesize that the zero-shot nature of these tasks can benefit from leveraging \emph{external knowledge} corresponding to the scene text.

In particular, we focus on the \mbox{Text-VQA} task, where understanding the scene text is necessary to answer questions about an image. The \emph{zero-shot} nature makes it susceptible to errors due to \emph{training data bias} when it fails to give due consideration to an instance only scene text cue and instead focuses more on trained statistics. E.g., in Figure \ref{fig:vsem_main}, the standard \mbox{Text-VQA} model incorrectly predicts the answer to be `apple' because of the computer-related visual content of the image. This observation motivates us to \emph{empower} scene text, like `commodore,' with \emph{context appropriate} valid knowledge. With the use of knowledge, we bring in contextual meaning about the unseen text `commodore' as a computer company which finally leads to the correct answer. We demonstrate more such cases in Figure \ref{fig:knog}, row 2. While one can mitigate such dataset biases by additional training on large annotated datasets, we argue for the use of \emph{external knowledge}.

The use of knowledge in Question Answering tasks has been proposed in the past but mainly applied to task-specific knowledge annotated datasets~\cite{shah2019kvqa, singh2019strings, mishraocr, Narasimhan_2018_ECCV, Wang2018FVQAFV}. We propose a more generalized way to add knowledge to any existing system. We leverage the availability of web scraped external knowledge base and apply it to existing \mbox{Text-VQA} datasets. Instead of relying on the availability of annotated knowledge facts, we extract multiple noisy candidate knowledge facts, which we validate through context before using. Furthermore, most works exploring the use of knowledge in VQA are set up to solve a \emph{retrieval problem}~\cite{shah2019kvqa, Narasimhan_2018_ECCV, Wang2018FVQAFV}, where the answer is retrieved as the most relevant knowledge fact. The answers in \mbox{Text-VQA} are more diverse, extending beyond the retrieved knowledge facts. Thus we deal with noisy knowledge facts and produce answers by reasoning with them. We address the more general problem of what is valid and relevant knowledge and how to reason with it to generate the answer. 

In Figure \ref{fig:vsem_main}, we present our core idea, which is to use external knowledge along with visual and scene text cues to answer questions about an image. We propose an end-to-end knowledge processing pipeline that extracts and validates web scraped knowledge and reasons with it to generate the answer.
We use the image context to filter out invalid candidate knowledge facts. Similar to the multimodal co-attention scheme of~\cite{ye2021breaking}, we propose a knowledge-enabled multimodal transformer~\cite{vaswani2017attention} to define task-specific relevance and reasoning with the validated knowledge facts.

Experimental results show that the integration of knowledge in our framework leads to improved results with respect to the baseline in generic \mbox{Text-VQA} datasets, reducing errors due to data bias. 
However, \mbox{Text-VQA} is primarily a reading task, and it does not always require knowledge to answer questions. Thus, to validate the efficacy of our scheme in dealing with knowledge, we also apply our model to \mbox{Text-KVQA}~\citep{singh2019strings}, a knowledge-enabled dataset along the lines of \mbox{Text-VQA}, with \emph{knowledge-oriented} questions \emph{designed to require knowledge} to answer. Our results highlight how our approach of using mined external knowledge leads to better results than using the associated ground-truth knowledge facts included with the \mbox{Text-KVQA} dataset.

\vfill\null
\vspace{-5pt}
In summary, our primary contributions are:
 \begin{itemize}
\item We present a method to extract external knowledge facts corresponding to scene text words in images and then validate the mined knowledge facts through image context. This validation stage helps us deal with the noisy nature of web scraped data.
\item The proposed end-to-end trainable architecture also enables us to reason with external knowledge. Our \emph{External Knowledge}-enabled \mbox{Text-VQA} \mbox{(\textbf{EKTVQA})} framework integrates the knowledge channel with a transformer-based architecture for the \mbox{Text-VQA} task. We propose the use of masked attention maps to control the interaction between the image components and the knowledge facts through the transformer layers. 
\item We apply our framework to standard \mbox{Text-VQA} datasets, demonstrating the effectiveness of leveraging external knowledge in improving the results. 
To the best of our knowledge, we are the first to propose the generalized use of external knowledge to existing \mbox{Text-VQA} datasets, where we show that some questions can be answered correctly only by retrieving and reasoning with external knowledge. Our framework gives results that match the state-of-the-art models without using additional annotated data or pre-training tasks. We further demonstrate the effectiveness of our model in answering the \emph{knowledge-oriented} questions of Text-KVQA to give state-of-the-art results.
\end{itemize}

The remainder of this paper is organized as follows. In Section~\ref{sec:relwork} we present the conventional approach to \mbox{Text-VQA} and relate it to existing knowledge-based systems. In Section~\ref{sec:KTM} we present our proposed method in which we incorporate external knowledge into the \mbox{Text-VQA} task. In Section~\ref{sec:exp} we report our experimental results and draw our insights, and finally, in Section~\ref{sec:concl} we present our concluding remarks.

\section{Related Works}\label{sec:relwork}
In this section, we first describe the \mbox{Text-VQA} task and the conventional solutions thus far applied to it. We then discuss the application of knowledge to Vision Language tasks and how it can be used to address the challenges in the \mbox{Text-VQA}. 

\subsection{\textup{\textbf{Conventional methods on \mbox{Text-VQA} task}}}\label{subsec:mmt}
The \mbox{Text-VQA} task has received widespread attention, both in the number of datasets~\cite{tmg_biten2019scene,tmg_singh2019textvqa,mishraocr} and network architectures~\cite{M4C_CVPR_2020,kant2020spatially,gao2020structured} proposed in the recent past. \mbox{Text-VQA} has two primary challenges -- the dynamic, open dictionary answer space and the multimodal nature of the task. 

\vspace{-5pt}
\subsubsection{\noindent{\textup{\textbf{Answer Space}}}}
The dynamic, open dictionary answer space implies that only around 60\% of the questions are answerable through a fixed vocabulary-based classifier of a traditional VQA setup. 
Therefore the initial baselines proposed were based on augmenting existing VQA architectures with OCR tokens and the formulation of a pointer network~\cite{tmg_singh2019textvqa} which allows predicting answer words from this OCR space~\cite{tmg_biten2019scene,tmg_singh2019textvqa,mishraocr}. However, the long-tailed distribution of the scene text implies that we often encounter previously unseen text blocks. Thus, while in theory, the pointer network can select the unknown scene text words, in practice, a vital piece of OCR token can get ignored during the answer prediction by virtue of being unseen or rarely seen in a similar image context. 
\vspace{-5pt}
\subsubsection{\noindent{\textup{\textbf{Multimodal}}}}
The second challenge is that \mbox{Text-VQA} is multimodal by definition. The Multimodal MultiCopy Mesh (\textbf{M4C})~\cite{M4C_CVPR_2020} model was proposed to leverage the attentional framework of the recently introduced transformer~\cite{vaswani2017attention} framework and address the multimodal nature of the data. The \textbf{multimodal transformer} based M4C jointly encodes the question, the image, and the scene text and employs an iterative decoding mechanism that supports multi-step answer decoding. The performance improvements attributed to M4C have led to multiple variants of this multimodal transformer-based architecture adopted in subsequent works~\cite{gao2020structured,kant2020spatially,yang2020tap}.

\vspace{-5pt}
\subsubsection{{\textup{\textbf{Architectural Updates and Pre-training Tasks}}}}
While M4C considers these text and visual objects alike and relies entirely on the internal transformer attentional framework, Structured Multimodal Attention (\textbf{SMA})~\cite{gao2020structured} has shown that a more structured attention, aware of the inter-object relationship, can lead to better results.

In Spatially Aware Multimodal Transformer (\textbf{SA-M4C})~\cite{kant2020spatially} the authors document that around 13\% of the questions have one or more spatial prepositions (e.g. `right,' `top,'`contains'). While \textbf{SA-M4C}~\cite{kant2020spatially} proposes encoding spatial relationships between detected objects through additional spatially aware self-attention layers atop M4C, in Text-Aware Pre-training \textbf{TAP}~\cite{yang2020tap} the authors achieve a similar objective through a pre-training task called Relative (spatial) Position Prediction (\textbf{RPP}). 

Besides the spatial reference to objects, around 20\% of the questions have explicit references to the OCR tokens. This \emph{OCR reference} artifact traces its roots to the dataset design, where the annotators were instructed to frame questions with the OCR tokens in consideration. 
For example, in Figure \ref{fig:knog}, row 3, column 2), the question refers to the book title and asks about the author (`who wrote afterlife ?'), instead of simply asking, `who wrote the book?'.
The Masked Language Modelling (\textbf{MLM}) proposed in \textbf{TAP}~\cite{yang2020tap} exploits these references to OCR tokens by pre-training to predict words from their immediate neighborhood. 

\begin{table*}[t]
\small
 \caption{External knowledge related work summary
 }
 \label{tab:relknw_sum}
 \setlength\tabcolsep{3 pt}
 \begin{tabular}{|l|l|l|l|l|l|}\hline
 &Work & Task &Dataset &Knowledge Source & Answering \\ \hline
 \rownumber & Marino et al. \citep{Marino2019OKVQAAV} & \makecell[l]{VQA task with \\knowledge-oriented \\questions} & \makecell[l]{Proposes a subset of\\ MSCOCO~\citep{lin2014microsoft} images\\ annotated with questions\\ that require external \\ knowledge} & \makecell[l]{Question words and scene\\ labels are used, as queries, \\ to retrieve articles from\\ Wikipedia} & \makecell[l]{Relevant sentences are identified \\
 from the retrieved articles based on \\
 the frequency of query words present \\
 and the top scoring word from the \\
 sentence is predicted as answer} \\\hline
 
 \rownumber & Li et al.\citep{Li2020BoostingVQ} & \makecell[l]{VQA task with \\knowledge-oriented \\questions} & \makecell[l]{Applied to existing \\ OK-VQA~\citep{okvqa},\\ FVQA~\citep{Wang2018FVQAFV}} & \makecell[l]{Visual entities and\\ question words are used to\\ query Wikidata~\citep{vrandevcic2014wikidata} and \\ ConceptNet~\citep{speer2012representing} } & \makecell[l]{The knowledge augmented\\ question words and visual features\\ are fused and fed to a classifier \\for answer generation} \\ \hline
 
 \rownumber & Wu et al.\citep{wu2021multi} & \makecell[l]{VQA task with \\knowledge-oriented \\questions} & \makecell[l]{Applied to existing \\ OK-VQA~\citep{okvqa}} & \makecell[l]{ Question and candidate \\answers are used to query \\ Wikipedia, ConceptNet and \\ retrieve similar images from \\Google image search } & \makecell[l]{The retrieved knowledge is \\ used to validate and rank the \\ candidate answers} \\ \hline 
 
 \rownumber & Wu et al. \citep{wu2016ask} &VQA Task & \makecell[l]{Applied to existing \\ VQA datasets~\citep{antol2015vqa,ren2015image} } & \makecell[l]{Attribute labels predicted\\ from visual view used to\\ query DBpedia } & \makecell[l]{Knowledge facts, encoded through\\ Doc2vec, are aggregated and\\ fed to an LSTM network for \\answer generation } \\ \hline
 
 \rownumber & Ye et al.\citep{ye2021breaking} &\makecell[l]{ Image Caption \\Pairing} & \makecell[l]{ Applied to Advertisement \\ Images~\citep{ad_hussain2017automatic}, and their \\ captions} & \makecell[l]{Scene text used to\\ query DBpedia } & \makecell[l]{Image guided attention mechanism\\ controls the inclusion of knowledge\\ facts into the image representation\\ trained with triplet loss} \\ \hline

 \end{tabular}
 
\end{table*} 
\vspace{5pt}
While pre-training tasks discussed above lead to better spatial and contextual understanding, they do not necessarily bring in new information about the nature of the scene text object to help with the zero-shot nature of the task. In fact, in most cases, the gains are more pronounced when the pre-training is across multiple annotated datasets, thus highlighting the importance of having access to different data distributions.
We attribute the zero-shot nature to the long-tailed distribution of the scene text and thus focus on empowering the scene text objects with external knowledge.

\subsection{\textup{\textbf{Knowledge applied to Vision Language Understanding Tasks}}}
Before we begin our discussion on knowledge-related works, we distinguish between valid and relevant knowledge facts. We feel this distinction to be necessary to highlight the gap areas in the related work. \textbf{Valid Knowledge} is one that is correct or valid for the given image. We propose checking for knowledge validity as an essential sub-task for any system that uses noisy web scraped knowledge facts. \textbf{Relevant Knowledge}, on the other hand, is not only valid for the image but also relevant to the current task of question-answering. While validity is specific to the image itself, relevance is both image and task-specific. Thus for a knowledge fact to be deemed relevant, it must be checked for validity first.

\vspace{-5pt}
\subsubsection{ \textup{\textbf{Knowledge-enabled Datasets}}} 
While the use of knowledge in question answering tasks has been seen in the past, they are mostly proposed together with a knowledge-enabled dataset~\cite{shah2019kvqa, singh2019strings, mishraocr, Narasimhan_2018_ECCV, Wang2018FVQAFV,Garca2020KnowITVA}, where the knowledge facts are part of the ground truth annotation provided.
Given a set of valid ground truth knowledge facts, most methods~\cite{Narasimhan_2018_ECCV, Wang2018FVQAFV, Narasimhan2018OutOT} use contextual relevance with the question and image to select a subset of \emph{relevant knowledge} facts. Interestingly in~\cite{Garca2020KnowITVA} we see an example where the question and ground truth candidate answers are used to retrieve knowledge facts.

In these knowledge-enabled datasets, the image and the corresponding ground truth knowledge facts are related by design. Thus the methods sidestep the issue of knowledge correctness or validity with respect to the image. In KVQA~\cite{shah2019kvqa}, for example, the dataset is built using image search against a list of persons; thus, every visual cue or person in the image is mapped explicitly to a knowledge cue related to that person. Similarly, we see in~\cite{singh2019strings} a large-scale knowledge-enabled dataset, \mbox{Text-KVQA}, comprising images of brands, book covers, and film posters with their names acting as anchor entities in the associated knowledge base provided as part of the dataset. Thus, the accompanying knowledge is always valid by design for these knowledge-enabled datasets. Such knowledge facts provide extra supervision and forgo the need for checking for knowledge validity. The process of making annotations is, however, resource-intensive and dataset-specific. We propose a more generalized approach of using noisy but cheap, web scraped external knowledge and applying it to existing datasets. 

\vspace{-5pt}
\subsubsection{\textup{\textbf{External Knowledge applied to Vision Language}}}
In Table \ref{tab:relknw_sum} we give a brief overview of some of the related work~\cite{Marino2019OKVQAAV,Li2020BoostingVQ,wu2021multi,wu2016ask,ye2021breaking} that leverages publicly available external knowledge bases, instead of using expensive ground truth knowledge facts as we saw in the last section. 
We observe that primarily it is the question words~\cite{Marino2019OKVQAAV,Li2020BoostingVQ} along with visual cues in the form of scene labels~\cite{Marino2019OKVQAAV}, detected entities~\cite{Li2020BoostingVQ}, or predicted visual attributes~\cite{wu2016ask}, that are used to retrieve knowledge from external sources. Once retrieved, the knowledge facts are incorporated into the \mbox{\emph{answer generation}}~\cite{Marino2019OKVQAAV,Li2020BoostingVQ,wu2016ask} based on their contextual relevance.

It is important to note that these methods~\cite{Marino2019OKVQAAV,Li2020BoostingVQ,wu2016ask,ye2021breaking} do not explicitly ensure \emph{knowledge validity} while attempting to model relevance. This is despite the fact that both DBpedia \footnote{https://dbpedia.org} and Wikipedia offer multiple candidate knowledge facts for a given query. Thus by skipping the validity checks, these methods allow room for spurious knowledge facts, which can be incorrectly deemed as relevant to the question.
This noisy retrieved knowledge, when reasoned with, burdens the answer predictor resulting in only marginal overall improvements \cite{wu2016ask,Marino2019OKVQAAV}. 
Thus in \citep{wu2021multi}, the authors propose the use of knowledge for \emph{answer validation} instead of \emph{answer generation}. 
In particular, the authors fine-tune a state-of-the-art VQA model~\cite{lu202012} to generate candidate answers, which are re-ranked using the retrieved candidate answer-specific knowledge. 
However, the zero-shot nature of \mbox{Text-VQA} implies that an unseen scene text may not be predicted as a candidate answer. Thus, retrieving knowledge based on candidate answers may result in the omission of relevant knowledge facts. We propose to acquire knowledge for \emph{all} scene text tokens and take on the challenge of filtering out noisy incorrect knowledge facts. 

None of the above methods make use of the \emph{scene text} while retrieving knowledge facts for question answering tasks. In~\cite{ye2021breaking} we find an example of scene text-based mined external knowledge from DBpedia but applied to the image-caption pairing task for Advertisement images~\cite{ad_hussain2017automatic}. The questions in \mbox{Text-VQA} primarily focus on reading and reasoning about the scene text objects. Thus, we exclusively use the scene text objects as our knowledge queries.

Furthermore, most of these methods~\cite{Marino2019OKVQAAV,Li2020BoostingVQ,wu2021multi} are designed for knowledge-oriented questions, where knowledge facts are \emph{always} required to generate an answer. Closest to our work is the application~\cite{wu2016ask} of mined external knowledge to existing VQA~\cite{antol2015vqa} Dataset, where the questions are not constrained to be knowledge-oriented.

\vspace{-5pt}
\subsubsection{\textup{\textbf{Using Knowledge to generate answer}}}
The use of knowledge in generating answers is either based on \emph{retrieval}~\citep{Marino2019OKVQAAV,Narasimhan_2018_ECCV, Wang2018FVQAFV} or \emph{reasoning}~\citep{Li2020BoostingVQ,wu2016ask} as seen in the related works. 
For knowledge fact-based VQA systems~\cite{Narasimhan_2018_ECCV, Wang2018FVQAFV}, where the answer is from a closed set of image-specific retrieved knowledge facts, answering a question translates to searching from amongst a set of knowledge facts. However, when the answer space extends beyond knowledge facts and requires generic English words, names, and numbers, one must \emph{reason} with retrieved knowledge to generate an answer. For example in~\citep{Li2020BoostingVQ,wu2016ask}, we see reasoning-based models where the knowledge is fused with image features and fed to a classifier~\citep{Li2020BoostingVQ} or an LSTM~\cite{wu2016ask} to generate the answer.

Text-VQA has a hybrid answer space comprising detected scene text tokens and a vocabulary of common answer words and numbers, as shown in~\cite{tmg_singh2019textvqa}. This diverse answer space demands reasoning with the knowledge facts. Thus, instead of fusing knowledge with image features ~\citep{Li2020BoostingVQ,wu2016ask}, we propose reasoning with the distinct knowledge facts corresponding to the individual scene text tokens. This allows us to distinguish between scene text tokens based on their associated knowledge facts. 

\begin{figure*}
\begin{center}
\includegraphics[width=1\textwidth]{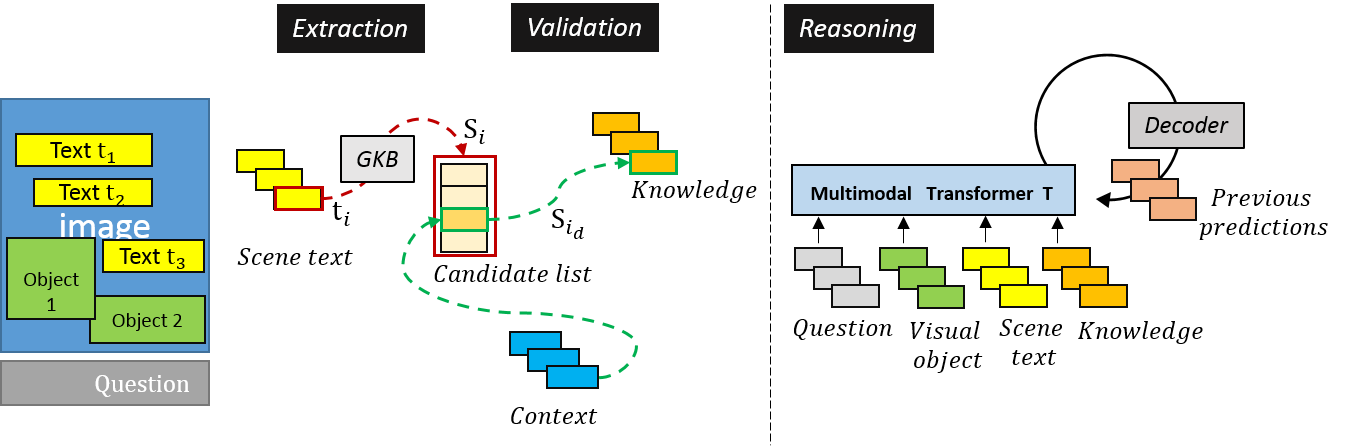}
\caption{ Extraction, Validation and Reasoning : The three stages of using external knowledge in \mbox{Text-VQA}. }\label{fig:ktm_dia}
\end{center}
\end{figure*}
\section{EKTVQA: External knowledge-enabled \mbox{Text-VQA}}\label{sec:KTM}

We now present our external knowledge-enabled \mbox{Text-VQA} framework, which proposes using a pre-existing external knowledge base to improve upon the \mbox{Text-VQA} task. \mbox{EKTVQA}, as shown in Figure \ref{fig:ktm_dia}, entails extracting, \mbox{validating}, and reasoning with noisy external knowledge in a multimodal transformer framework.
The input to the \mbox{Text-VQA} task is a question-image pair. 
Thus the input stream consists of question words along with detected visual objects and scene text. To augment our understanding of the scene text, we use an external knowledge base to extract candidate meanings, which we validate and filter through context. 
Similar to existing works~\cite{M4C_CVPR_2020,kant2020spatially,yang2020tap}, we build upon a multimodal transformer-based architecture. We incorporate knowledge into the framework through the inclusion of a separate knowledge channel and achieve constrained interaction through attention masks.

We start by first presenting our external knowledge extraction and validation stages, followed by details about the different feature formulations used, and finally discuss reasoning with external knowledge to generate answers. 

\begin{figure*}
\begin{center}
\includegraphics[width=0.95\textwidth]{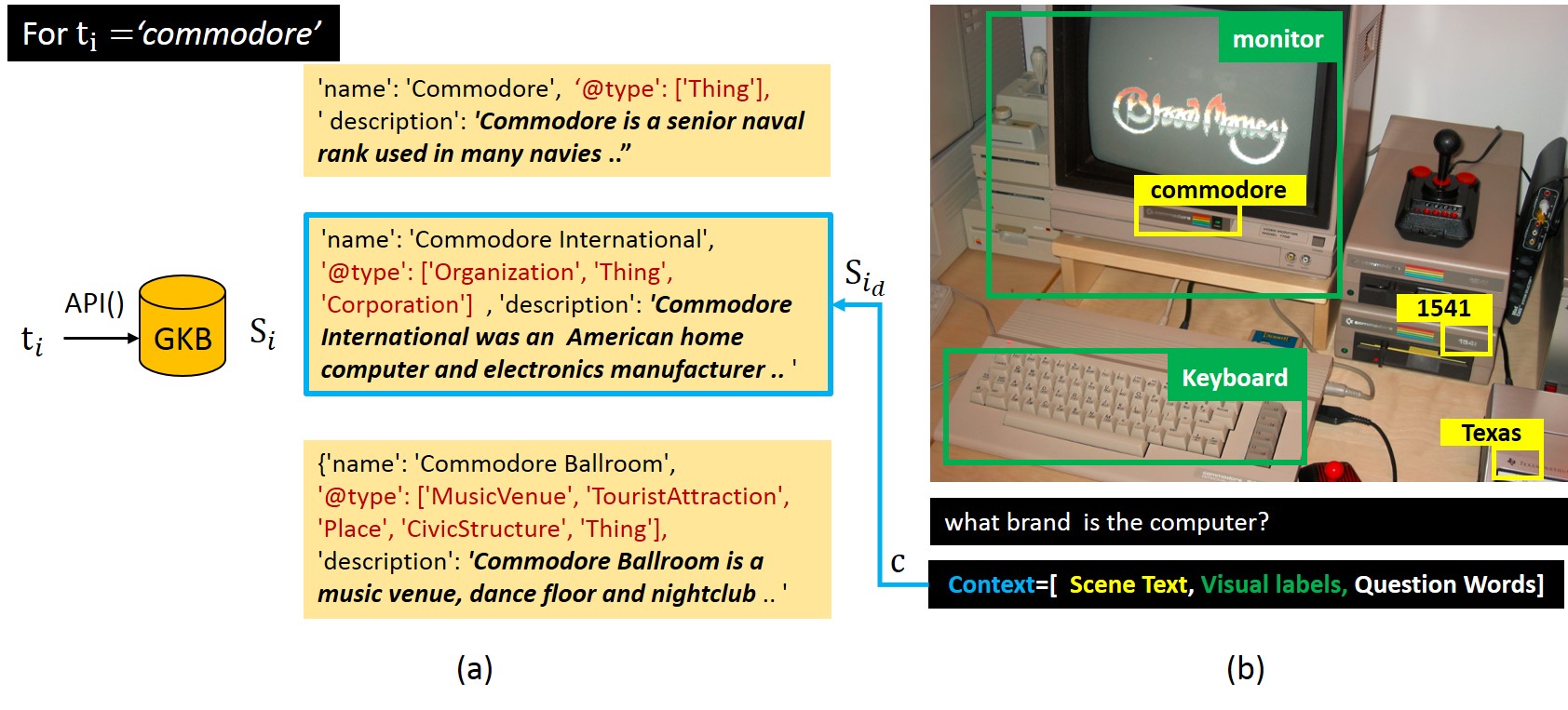}
\vspace{-15pt}
\caption{ Knowledge Extraction and Validation. Figure (a) shows knowledge extraction using Google API and Figure (b) demonstrates knowledge validation and selection through image context. We identify knowledge validation as an important subtask for web scraped knowledge data that is not anchored to our \mbox{Text-VQA} datasets explicitly}\label{fig:kno_ext_sel}
\end{center}
\end{figure*}

\subsection{\textup{\textbf{Knowledge Extraction And Validation}}}\label{para:knwextvld}
\subsubsection{\textup{\textbf{Knowledge Extraction from External Source }}}\label{para:knwext}
For each detected scene text word, we query Google Knowledge Base (GKB) \footnote{https://developers.google.com/knowledge-graph}. In case of a hit in GKB, for each query scene text word $\mathbf{t}_i$, we get a set of candidate meanings $\mathbf{S}_i$, which we retain and later disambiguate in the knowledge validation step.
In Figure \ref{fig:kno_ext_sel}a, we show a sample response we get. With the default GKB API setting of a maximum of 10 candidates per query, we observe that the median number of candidates returned per query word is 4 for a validation set of 25000 words. Based on this observation, we limit the maximum number of candidates to 4. 

\vspace{-5pt}
\subsubsection{\textbf{Knowledge Validation through Context}}\label{para:knwflt} 
The particular meaning a scene text word assumes depends on the context it occurs in. Thus disambiguation amongst candidate meanings is done based on the context of the given image. We see an example of this in Figure \ref{fig:kno_ext_sel}b, where the context words like `monitor' and `keyboard' select the valid knowledge description for `commodore.'

We define the image context $\mathbf{c}=[\mathbf{t},\mathbf{vc},\mathbf{qw}]$ as the aggregation
of 3 types of entities $-$ OCR tokens $\mathbf{t}$, visual object labels $\mathbf{vc}$ and
question words $\mathbf{qw}$. We consider \textit{N}, \textit{V}, \textit{L} as the counts for the OCR tokens,
object labels and question words, respectively. 
This context is encoded in BERT~\cite{Devlin2019BERTPO} to generate $\{\mathbf{x}^{\text{c}}_{p}\}$ with $\mathbf{x}^{\text{c}}_p$ $\in$ $R^{768}$ (where \textit{p} = 1, . . . , \textit{N+V+L}).

Given a scene text word $\mathbf{t}_i$ and its set of candidate knowledge facts $\mathbf{S}_i=\{\mathbf{S}_{i_{j}}\}$, we use the context $\mathbf{c}$ of the current image to select the valid knowledge $\mathbf{S}_{i_{d}} \in \mathbf{S}_i$. 
For each candidate knowledge $ \mathbf{S}_{i_{j}} \in \mathbf{S}_i$, we learn a validity score $\mathbf{r}_j$, given by Eq. \ref{eqn:rel1} which compares its BERT embedding with those of the context words. Using Eq. \ref{eqn:rel2} and Eq. \ref{eqn:rel3} we deem the highest scoring candidate knowledge as being image context-appropriate, therefore, valid. 
The BERT embedding of this valid knowledge fact corresponding to $\mathbf{t}_i$ is referred to as $\mathbf{x}^{\text{knw}}_{i}$. 
\vspace{-3pt}
\begin{align}\label{eqn:rel1}
& \mathbf{r}_j=\mathbf{W_c}( Relu(\mathbf{W_a} \mathbf{x}^{\text{S}_{i_{j}}}) \cdot Relu(\mathbf{W_a} \sum \mathbf{x}^{\text{c}}_p ) )
\end{align}
\begin{align}\label{eqn:rel2}
& d=\argmax_j \mathbf{r}_j 
\end{align}
\begin{align}\label{eqn:rel3}
& \mathbf{x}^{\text{knw}}_i=\mathbf{x}^{\text{S}_{i_{d}}}
\end{align}

\noindent{Here $\mathbf{W_a,W_c}$, are learned projection matrix. Further, $\mathbf{x}^{\text{S}_{i_{j}}}$ and $\mathbf{x}^{\text{c}}$, are the BERT embeddings of $j^{\text{th}}$ candidate meaning $\mathbf{S}_{i_{j}}$ and context $\mathbf{c}$ respectively. Our implementation uses a softmax, retaining only the candidate knowledge with the highest probability.}

\subsection{\textup{\textbf{Multimodal Features}}}\label{subsec:modelfeatures}The input to our multimodal transformer, as illustrated in both Figures \ref{fig:ktm_dia} and \ref{fig:ktm_main}, consists of the given question, detected visual objects and scene text, validated external knowledge facts, and previous decoding states. 
\vspace{-5pt}
\subsubsection{\textup{\textbf{Question Words}}} Given a sequence of \textit{L} question words, we embed them as $\{ \mathbf{x}_l^{\text{ques}} \}$ (where \textit{l} = 1, . . . , \textit{L}) using pre-trained BERT~\cite{Devlin2019BERTPO} with $\mathbf{x}_l^{\text{ques}}$ $\in$ $R^{768}$.

\vspace{-5pt}
\subsubsection{\textup{\textbf{Visual Objects}}} Following M4C, each of the \textit{M} visual object regions are encoded as $\mathbf{x}^{\text{fr}}_m$, using a pre-trained ResNet based Faster R-CNN (\textbf{FRCNN})~\cite{ren2015faster} model trained on Visual Genome. Additionally the bounding box information is embedded as $\mathbf{x}^{\text{b}}_m$ to preserve spatial information. The appearance and spatial features thus obtained are linearly projected and combined to generate the final set of visual object features $\{\mathbf{x}^{\text{obj}}_m\}$ (where \textit{m} = 1, . . . , \textit{M}) as 
\begin{equation}\label{feat:obj}
\mathbf{x}^{\text{obj}}_m= LN(\mathbf{W_1} \mathbf{x}^{\text{fr}}_m) + LN(\mathbf{W_2} \mathbf{x}^{\text{b}}_m) \\
\end{equation}
\noindent{where $\mathbf{W_1}$ and $\mathbf{W_2}$ are learned projection matrices and $LN()$
denotes layer normalization.}

\vspace{-5pt}

\subsubsection{\textup{\textbf{Scene Text}}} For the \textit{N} OCR tokens, we encode their visual appearance, recognized text, and spatial information. The same FRCNN based detector, discussed above, extracts and encodes the visual appearance feature corresponding to the bounding box as $\mathbf{x}^{\text{fr}}_n$. This is combined with the Fasttext~\cite{fasttext} and PHOC~\cite{phoc} embeddings of the OCR token, namely $\mathbf{x}^{\text{ft}}_n$ and $\mathbf{x}^{\text{ph}}_n$ respectively.
Similar to the visual objects, the bounding box coordinates are embedded as $\mathbf{x}^{\text{b}}_n$ to encode the spatial information. Finally, we define the scene text features $\{\mathbf{x}^{\text{ocr}}_n\}$ (where \textit{n} = 1, . . . , \textit{N}) as
\begin{equation}\label{feat:ocr}
\mathbf{x}^{\text{ocr}}_n= LN( \mathbf{W_3} \mathbf{x}^{\text{ft}}_n + \mathbf{W_4} \mathbf{x}^{\text{fr}}_n + \mathbf{W_5} \mathbf{x}^{\text{ph}}_n ) + LN(\mathbf{W_6} \mathbf{x}^{\text{b}}_n)\\ 
\end{equation}
\noindent{where $\mathbf{W_3}$,$\mathbf{W_4}$, $\mathbf{W_5}$ and $\mathbf{W_6}$, are learned matrices.}

\vspace{-5pt}
\subsubsection{\textup{\textbf{External Knowledge}}} Knowledge obtained from Google Knowledge Base (\textbf{GKB}) is in the form of a text description, which is encoded using pre-trained BERT sentence embedding. Corresponding to each encoded OCR token $\mathbf{x}^{\text{ocr}}_n$, we have a knowledge node $\mathbf{x}^{\text{knw}}_{n}$ $\in$ $R^{768}$. This knowledge node is selected from among the candidate meanings following the procedure explained earlier in Section~\ref{para:knwextvld}.

\vspace{-5pt}
\subsubsection{\textup{\textbf{Previous Prediction}}} The answer is generated step-wise, one word at a time, either from the vocabulary of common answer words or current image specific OCR tokens. As seen in Figure \ref{fig:ktm_main}, the current step answer word prediction requires, as part of its input, the predicted answer word from the previous \textit{D} steps $\{\mathbf{x}^{\text{prv}}_d$\} (where \textit{d} = 1, . . . , \textit{D}) and $\mathbf{x}^{\text{prv}}_d$ $\in$ $R^{768}$.

If at step $d-1$ predicted answer word is the $n^{\text{th}}$ OCR token, we feed its OCR representation $\mathbf{x}^{\text{ocr}}_n$ as the input $\mathbf{x}^{\text{prv}}_d$ for the current step $d$. Otherwise, if the previous step prediction is the $i^{\text{th}}$ vocabulary word, we set $\mathbf{x}^{\text{prv}}_d$ to $\mathbf{W}_{i}^{\text{voc}}$, the $i^{\text{th}}$ column of the learned projection matrix $\mathbf{W}^{\text{voc}}$ discussed in Section \ref{subsec:ansgen}.
For a more detailed discussion, we refer the reader to ~\cite{M4C_CVPR_2020}.

\subsection{\textup{\textbf{Reasoning with External Knowledge}}}\label{subsec:modeladj} 
Reasoning with external knowledge implies identifying relevant knowledge facts with respect to question and image context and linking them to their corresponding detected OCR tokens. We now discuss how \textbf{relevance} is achieved through transformers, followed by our proposed \textbf{constrained interaction} which incorporates knowledge into scene text representation while ensuring legitimacy, and finally present our \textbf{answer generation} scheme. 
\vspace{-5pt}
\subsubsection{\noindent\textbf{Relevance}} While the validity or correctness of the external knowledge can be judged from its image context, its relevance or applicability while generating answers remains to be ascertained.
 The decision to include or exclude the external knowledge fact about a scene text word is crucial when it comes to generating answers. Not every question requires external knowledge to be answered. While the answer may still comprise scene text words, we do not always require the corresponding knowledge fact about the scene text word. In contrast to datasets like~\cite{okvqa,shah2019kvqa} that explicitly require knowledge, \mbox{Text-VQA} provides a more realistic setup, and thus the decision to use the knowledge fact is contingent on it being relevant. Instead of an explicit task to define knowledge relevance like~\cite{Narasimhan_2018_ECCV, Wang2018FVQAFV}, we include the set of knowledge nodes $\{\mathbf{x}^{\text{knw}}_{n}\}$ as part of the transformer input and let the internal attentional framework~\cite{vaswani2017attention} select relevant knowledge facts. The transformer attentional framework treats the scene text, visual objects, question words, and knowledge nodes alike, attending to relevant nodes based on the final task of question answering.

\vspace{-10pt}
\subsubsection{\noindent \textbf{Constrained Interaction} } The self-attention layer~\cite{vaswani2017attention} in the transformer architecture allows a node to form its representation using the other nodes in the neighbourhood. In practice, this neighborhood is typically the entire set of co-occurring nodes, but it can also be constrained to a smaller subset, thereby limiting which nodes interact. In the past, such constrained interaction through the use of attention masks has found application in causal step-wise decoding~\cite{M4C_CVPR_2020} and enforcing spatial relations~\cite{kant2020spatially}. In this work, we extend the use of constrained interaction to preserve \textbf{node-to-node knowledge legitimacy} i.e., the one-to-one relationship between the scene text nodes and their corresponding validated knowledge facts. Instead of simply fusing the scene text token and the corresponding knowledge fact into a single entity, we let them interact with the question and image context and influence the final scene text representation. 

\begin{figure}
\begin{center}
\includegraphics[width=0.4\textwidth]{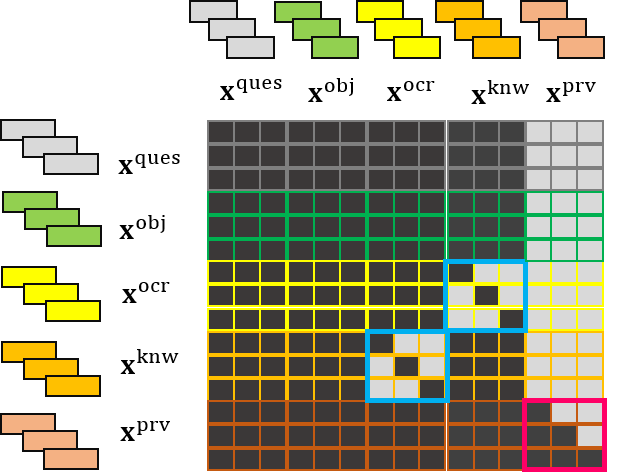}
\vspace{-5pt}
\caption{ Constrained Interaction: Knowledge inclusion through Attention Mask matrix A. It controls the interaction between the nodes. Dark Cells allow interaction between corresponding nodes. $\mathbf{A}^{KT}$ and $\mathbf{A}^{TK}$, highlighted in blue, are $N \times N$ identity matrices, binding the knowledge facts with their corresponding OCR tokens. Highlighted in pink is the $D \times D$ lower triangular sub-matrix $\mathbf{A}^{PP}$ enforcing causality between previous decoding steps }\label{fig:kno_adj}
\end{center}
\end{figure}

\noindent Formally, the Attention Mask matrix $\mathbf{A} \in R^{E \times E}$ controls the interaction between the $L$ question words, $M$ visual objects, $N$ OCR objects, $N$ knowledge nodes, and $D$ previous stage predictions, with the total number of interacting elements as $E=L+M+N+N+D$. Specifically, $\mathbf{A}_{ij}$ is added as a bias term to the softmax layer of the self-attention, effectively controlling the $j^{\text{th}}$ element's role in the $i^{\text{th}}$ element's new representation. 
For intuitive understanding, the dark cells in Figure \ref{fig:kno_adj} represent 0, and the grey/empty cells represent $-\infty$. As seen in the figure, our Attention Mask matrix $\mathbf{A}$ allows complete interaction between the visual, scene text, and question words. However, we note that the interaction between the scene text words $\{\mathbf{x}^{\text{ocr}}_{n}\}$ and their corresponding knowledge facts $\{\mathbf{x}^{\text{knw}}_{n}\}$ is subject to certain constraints. 

\begin{figure*}
\begin{center}
\includegraphics[width=1\textwidth]{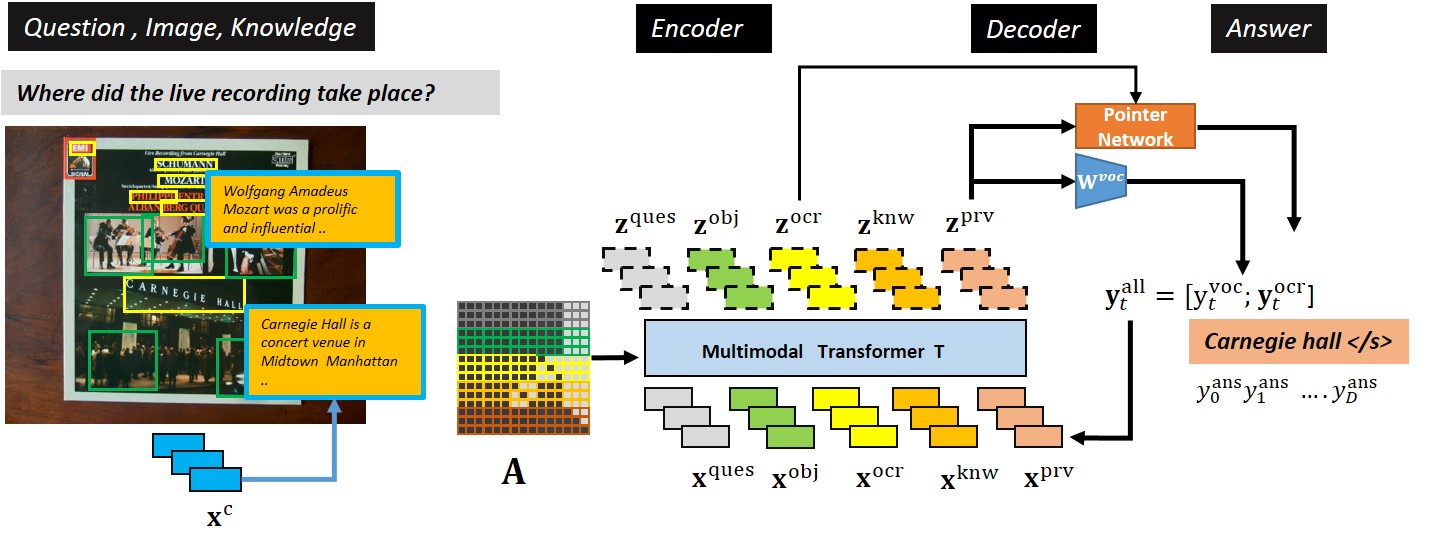}
\caption{ EKTVQA: Our proposed External Knowledge-enabled \mbox{Text-VQA} }\label{fig:ktm_main}
\end{center}
\end{figure*}
\noindent The scene text words are \textbf{symbolic} representations, with the mined knowledge facts bringing in the associated \textbf{semantics}. This symbolic-semantic one-to-one relation needs to be preserved if we want to later discriminate amongst scene text words based on knowledge. 
As explained in Section~\ref{subsec:ansgen} answer generation entails comparing the transformer output $\{\mathbf{z}^{\text{ocr}}_{n}\}$ of the scene text words with the current decoding state $\mathbf{z}^{\text{prv}}_{d}$. 
Thus, we must ensure that for a given scene text word $\mathbf{x}^{\text{ocr}}_{n}$, only its own legitimate knowledge fact $\mathbf{x}^{\text{knw}}_{n}$ contributes to this final representation $\mathbf{z}^{\text{ocr}}_{n}$.
This is enforced through the identity sub-matrices $\mathbf{A}^{KT},\mathbf{A}^{TK}$, highlighted in Figure \ref{fig:kno_adj}, preserving \textbf{node-to-node knowledge legitimacy} by allowing only the related knowledge fact and OCR token to attend to each other. Further, the lower triangular sub-matrix $\mathbf{A}^{PP}$ enables step-wise causal decoding~\cite{M4C_CVPR_2020,kant2020spatially}, enforcing that the current answer generation can only interact with previously decoded answer tokens.

\noindent Formally our multimodal transformer $T$ can be defined as, 
\begin{equation}\label{feat:tr1}
[\mathbf{z}^{\text{q}},\mathbf{z}^{\text{obj}},\mathbf{z}^{\text{ocr}},\mathbf{z}^{\text{knw}},\mathbf{z}^{\text{prv}}]= T([\mathbf{x}^{\text{q}},\mathbf{x}^{\text{obj}},\mathbf{x}^{\text{ocr}},\mathbf{x}^{\text{knw}},\mathbf{x}^{\text{prv}}],\mathbf{A})
\end{equation}

\vspace{-5pt}
\subsubsection{\textup{\textbf{Answer Generation: }}}\label{subsec:ansgen} Answer Generation consists of stepwise causal decoding of the transformer outputs, akin to the iterative decoder introduced in~\cite{M4C_CVPR_2020}. We generate the answer tokens $y^{\text{ans}}_d, \forall d$, one token at a time, from the current decoding state, predicting either from the fixed vocabulary of $H$ words or from amongst the current $N$ OCR tokens.

The current decoding state $\mathbf{z}^{\text{prv}}_{d}$ is linearly projected through $\mathbf{W}^{\text{voc}}$ to generate the \textit{H} dimensional vocabulary scores $\mathbf{y}^{\text{voc}}_{d}$. The \textit{N} dimensional OCR scores $\mathbf{y}^{\text{ocr}}_{d}$ are generated through a bi-linear interaction between $\mathbf{z}^{\text{prv}}_{d}$ and the OCR outputs $\{\mathbf{z}^\text{{ocr}}_n\}$ (where \textit{n} = 1, . . . , \textit{N}). We refer to this as the Pointer network in Figure \ref{fig:ktm_main}. We take the argmax on the concatenated score $\mathbf{y}^{\text{all}}_d=[\mathbf{y}^{\text{voc}}_{d},\mathbf{y}^{\text{ocr}}_{d}]$ to generate the current answer token $y^{\text{ans}}_d$. We successively decode for 12 steps or till we predict the special end marker token.


\section{Experiments and Results}\label{sec:exp}
We benchmark our method on the \mbox{TextVQA} dataset~\cite{tmg_singh2019textvqa} , the ST-VQA~\cite{tmg_biten2019scene} dataset and the Text-KVQA dataset~\cite{singh2019strings}. We briefly present our implementation and dataset details, followed by comparison, analysis, ablation study, limitations, and finally, a summary of our results.

\subsection{\textup{\textbf{Implementation and Dataset details}}}\label{subsec:datasets}

\subsubsection{\textup{\textbf{{Implementation Details}}}}
Our implementation is based on PyTorch Transformers library \footnote{https://pytorch.org/hub/huggingface\_pytorch-transformers}. 
We adopt a multimodal transformer architecture akin to~\cite{M4C_CVPR_2020}. Our basic transformer design is similar in spirit, with eight attention heads and four layers, but we add masking to constrain the external knowledge interaction. We use Adam as our optimizer. We start with a learning rate of $1e-4$, reducing it step-wise upon stagnation or plateau.

\subsubsection{\textup{\textbf{{Dataset Details}}}}
\paragraph{\textup{\textbf{Dataset 1: \mbox{TextVQA}}}} The \mbox{TextVQA} dataset consists of 28,408 images collected from OpenImages~\cite{Kuznetsova_2020}, with 45,336 question-answer pairs. 
The proposed evaluation metric is accuracy, measured via soft voting of the ten ground-truth answers.

We use Google-OCR \footnote{https://cloud.google.com/vision/docs/ocr}, for scene text detection and recognition. Of the total 189551 detected scene text words, GKB offers an average of 2.29 different candidate meanings for each of the 100972 words found in the Knowledge Base. 
In Figure \ref{fig:kno_ext_sel}a, we see that the GKB response includes 3 tags $-$ name, description and attribute.
Interestingly we note that GKB exhibits tolerance towards spelling errors, often returning candidates with high lexical similarity. While in some cases, this leads to correct candidate knowledge facts even with incorrect OCR, it also leads to a significant number of false positives. 
Therefore we remove entries where the queried scene text is not present in the `name' or `description' tag of the returned candidate meaning. For the remaining words, we merge their description and attribute tags. This results in mined knowledge facts for 57666 words, including 2917 multiword entity names (e.g., `new york,' `burger king').

\begin{table*}[h]
\setcounter{magicrownumbers}{0} 
 \begin{center}
\small
\caption{Results of \mbox{Text-VQA} Task on \mbox{TextVQA} and ST-VQA Dataset}\label{tab:sotacomp}
 \begin{tabular}{|l|l|c|c |c|c|c|c|c|c }\hline
& & \multicolumn{3}{|c|}{TextVQA Dataset} & \multicolumn{3}{|c|}{ST-VQA Dataset} \\ \hline 
 &Model & OCR &Val. Acc. &Test Acc. & OCR &Val. ANLS. &Test ANLS \\ \hline
\rownumber&LoRRA \cite{tmg_singh2019textvqa} & Rosetta-ml & 26.56 &27.63 & - &- &- \\
\rownumber&M4C \cite{M4C_CVPR_2020} & Rosetta-en & 39.44 &39.10 & Rosetta-en & 0.472 &0.462 \\ 
\rownumber&SMA \cite{gao2020structured} & Rosetta-en & 40.05 &40.66 & Rosetta-en & - &0.466 \\
\rownumber&CRN \cite{crn} & Rosetta-en & 40.39 &40.96 & Rosetta-en & - &0.483 \\
\rownumber&LaAP-Net \cite{han2020finding} & Rosetta-en & 40.68 &40.54 & Rosetta-en & 0.497 &0.485 \\
\rownumber&SA-M4C \cite{kant2020spatially} & Google-OCR & 43.9 & - & Google-OCR & 0.512 &0.504 \\ 
\rowcolor{lightgray} \rownumber&TVQA & Google-OCR & 41.7 & - & Rosetta-en & 0.468 &-\\
\rowcolor{GreenYellow} \rownumber&EKTVQA & Google-OCR & 44.26 & 44.2 & Rosetta-en & 0.499 & 0.48\\ \hline 
\end{tabular}
\end{center} 
\end{table*}

\begin{table*}[h]
\setcounter{magicrownumbers}{0} 
 \begin{center}
\small
\caption{Results of \mbox{Text-VQA} Task on Text-KVQA (Scene) Dataset. Results marked with * do not use the exactly our same partitions for training and test.}\label{tab:sotaEKTVQA}
 \begin{tabular}{|l|l|c|c|}\hline
 &Model & OCR &Test Acc. \\ \hline
\rownumber&TextKVQA \cite{singh2019strings} & Pixellink\cite{Deng2018PixelLinkDS} + CRNN\cite{Shi2017AnET} & 54.5* \\
\rownumber&TextKVQA \cite{singh2019strings} & Textspotter\cite{liao_mask_textspotter} & 52.4* \\
\rowcolor{lightgray} \rownumber&TVQA & Textspotter\cite{liao_mask_textspotter} & 75.4 \\
\rowcolor{LimeGreen} \rownumber&KBVQA & Textspotter\cite{liao_mask_textspotter} & 77.06 \\
\rowcolor{GreenYellow} \rownumber&EKTVQA & Textspotter \cite{liao_mask_textspotter} & 77.81 \\ \hline 
\end{tabular}
\end{center} 
\vspace{-5pt}
\end{table*}

\begin{table*}[h]
\setcounter{magicrownumbers}{0} 
 \begin{center}
\small
\vspace{-3pt}
\caption{Results of related methods on \mbox{TextVQA} Dataset with additional pre-training tasks and datasets}\label{tab:sotapre}
 \begin{tabular}{|l|l|c|c|c|c |c|c|c}\hline
 & & \multicolumn{2}{|c|}{Pre-training} & & \multicolumn{2}{|c|}{TextVQA Dataset} \\ \hline
 &Model &Pre-training Task & Pre-training Dataset & OCR &Val. Acc. &Test Acc. \\ \hline
\rowcolor{GreenYellow} \rownumber &EKTVQA &- &- &Google-OCR & 44.26 & 44.2 \\ 
\rownumber &M4C \cite{M4C_CVPR_2020} &- &ST-VQA & Rosetta-en & 40.55 &40.46 \\
\rownumber &SA-M4C \cite{kant2020spatially} &- &ST-VQA & Google-OCR & 45.4 & 44.6 \\
\rownumber &TAP \cite{yang2020tap} &MLM,ITM,RPP &- & Rosetta-en & 44.4 &- \\
\rownumber &TAP \cite{yang2020tap} &MLM,ITM,RPP &- & Microsoft-OCR & 49.91 &49.71 \\
\rownumber &TAP \cite{yang2020tap} &MLM,ITM,RPP &ST-VQA,Textcaps,OCR-CC & Microsoft-OCR & 54.71 & 53.97 \\\hline
 \end{tabular}
\end{center} 
\end{table*}
\vspace{-7pt}
\paragraph{\textup{\textbf{Dataset 2: ST-VQA}}}The ST-VQA dataset consists of 23,038 images, from various sources~\cite{icdarrr13,icdarrr15,Vizwiz,ImageRU,VisualGC,COCOTextDA}, with 31791 question-answer pairs. To deal with OCR recognition errors that lead to answers with misplaced /missing characters, the authors~\cite{tmg_biten2019scene} propose the Average Normalized Levenshtein Similarity (ANLS) as an alternate evaluation metric. Normalized Levenshtein Similarity (NLS) is defined as,
\begin{equation}\label{anls}
\textup{NLS}=1 - d_L(y^{\text{ans}},y^{\text{GT}})/max(y^{\text{ans}},y^{\text{GT}})
\vspace{-.5pt}
\end{equation}
\noindent{where $y^{\text{ans}}$ and $y^{\text{GT}}$ are the predicted and ground truth answers respectively, and $d_L$ refers to the edit distance. ANLS refers to averaged NLS scores over all questions, with scores below 0.5 being truncated to 0 before averaging.}
For scene text words, we use the Rosetta-en OCR tokens from Pythia~\cite{singh2018pythia}.
From the total of 64830 detected scene text words, GKB offers candidate meanings for 39053 words. The final knowledge annotations mined, after the filtering described above, consist of 23132 words and their candidate meanings.

\vspace{-7pt}
\paragraph{\textup{\textbf{Dataset 3: Text-KVQA(Scene)}}}
The Text-KVQA dataset consists of images of business brands harvested through Google Image Search and hyperlinks to film posters and book covers present at external sources. While the book and film images are poster-like, the business images are natural images of outdoor scenes pertaining to business brands, which, coupled with their ready availability by the authors~\citep{singh2019strings}, makes this partition our point of investigation. In summary, these business brand images referred to as the scene partition, consist of 10436 images and 64870 associated question-answer pairs. Since no exact train-test partition was provided, we train on a random partition of $80\%$ split of the images and test on the rest, similar to the authors~\citep{singh2019strings}. Unique to this dataset is the associated knowledge facts provided for each image. However, the knowledge facts do not explicitly correspond to any scene text. Thus, in tune with our existing framework, we extract candidate knowledge facts for 10952 of the 24361 detected scene text words using GKB.

\begin{table}[h]
 \begin{center}
\small
\caption{ Dataset Comparison. \textbf{OCR UB}: Percentage of answers comprising of OCR tokens alone, \textbf{Vocab UB}: Percentage of answers comprising of fixed vocabulary words only, \textbf{OCR + Vocab UB}: Percentage of answers comprising of OCR tokens and answer vocabulary words, \emph{i.e.}, the maximum achievable accuracy.}\label{tab:dscomp}
 \resizebox{\linewidth}{!}{\begin{tabular}{|l|c|c|c| }\hline
Dataset & \mbox{TextVQA} &ST-VQA &Text-KVQA \\ \hline
 \#Images & 28408 &23038 & 10436\\
\#Question-Answer & 45,336 &31791 & 64870\\ \hline 
OCR system & Google-OCR &Rosetta-en &TextSpotter\\
\#Scene-Text & 189551 &64830 & 24361\\
\#Mined Knowledge & 57666 &23132 & 10952\\ \hline
OCR UB (\%) & 53.4 & 50.53 & 5.76 \\
Vocab UB (\%) & 57.14 & 67.16 & 97.65\\ 
OCR + Vocab UB (\%) & 84.4 & 83.33 & 97.96\\ \hline

 \end{tabular}}
\end{center} 
\end{table}
\vspace{-5pt}
\subsection{\textup{\textbf{Comparison with state-of-the-art}}}\label{subsec:ressota}
In Table \ref{tab:sotacomp} we report our \mbox{Text-VQA} open-ended question answering results on the datasets \mbox{TextVQA} and ST-VQA. 
In row 7, we present our ablated instance without knowledge, namely TVQA, and in row 8, we present our full model EKTVQA. For all models, we specify the OCR system used on the dataset. 

Our primary observation is the $2.5\%$ absolute improvement of EKTVQA over non-knowledge baseline TVQA across datasets.
This is interesting when we consider the $0.92\%$ improvement obtained by~\citep{wu2016ask} over their non-knowledge baseline \textbf{Att+Cap+LSTM}~\citep{wu2016ask} when applied to the VQA dataset. For both these tasks, VQA and \mbox{Text-VQA}, the questions are not knowledge-oriented, but the improvements due to external knowledge are significantly more for \mbox{Text-VQA}. This can be attributed to the zero-shot characteristics of the \mbox{Text-VQA} task, which benefits from the external knowledge about scene text. 

For the \mbox{TextVQA} dataset, our model improves upon the state-of-the-art~\cite{kant2020spatially}. For the ST-VQA dataset, our results are comparable with models with a similar OCR system. Table \ref{tab:sotacomp} also highlights the correlation between a better OCR system and improved answer generation. We observe that, in general, models with the Google-OCR system perform better than those with Rosetta OCR systems. 
To elucidate this further, in Table \ref{tab:compocr}, we report the Upper Bound accuracies~(\textbf{UB}) on the \mbox{TextVQA} validation set against the different OCR systems. The OCR systems can be judged by comparing their OCR Upper Bound accuracy, which is the percentage of ground-truth answers that can be composed of OCR tokens alone. For the validation set, the Rosetta-en OCR tokens have an OCR Upper Bound accuracy of $44.9\%$ compared to $53.4\%$ of Google-OCR, as shown in Table \ref{tab:compocr}. 
For the ST-VQA dataset, we only have Rosetta-en OCR tokens, which, as we have just seen from the \mbox{TextVQA} dataset, are not as accurate as the Google-OCR tokens. Thus for the ST-VQA dataset, the erroneous Rosetta-en tokens results in not just missing out on answers words, leading to a lower OCR UB compared to \mbox{TextVQA}, but also amounts to fewer knowledge hits compared to \mbox{TextVQA}, as listed in Table \ref{tab:dscomp}.
Thus our underwhelming performance on the ST-VQA dataset, despite the use of knowledge, can be attributed to the erroneous Rosetta-en OCR tokens.

While our results on the \mbox{Text-VQA} task are encouraging, it is primarily a reading task that does not always require external knowledge to answer. Thus, we also present our results on the Text-KVQA dataset, which is explicitly designed to require knowledge to generate answers. 
In Table \ref{tab:sotaEKTVQA} we compare our generalized knowledge pipeline against ground truth knowledge on the Text-KVQA dataset. While \mbox{TextKVQA} is the knowledge-enabled memory network-based model proposed by the authors~\citep{singh2019strings}, we also present an ablated instance called KBVQA, which is an approximation to our full model EKTVQA, but uses the provided image-level knowledge facts instead of GKB queried knowledge facts corresponding to scene text tokens. It is important to note that in KBVQA, there is no knowledge validation phase as the knowledge used is part of the ground truth information. Further, as the knowledge facts do not correspond to scene text, we did not make use of constrained interaction to ensure knowledge legitimacy. Note that knowledge legitimacy, one of our core ideas in EKTVQA, helps us select legitimate scene text answer words corresponding to relevant knowledge facts.

For the Text-KVQA dataset, however, only about $6\%$ of the questions are answerable using scene text, compared to more than $50\%$ for both \mbox{TextVQA} and ST-VQA datasets, as is shown in Table \ref{tab:dscomp}. This low OCR UB explains why our full model EKTVQA, which includes constrained interaction, is not able to significantly improve upon the KBVQA ablation.
The high performance of our non-knowledge baseline TVQA begs a closer inspection. While the \mbox{Text-VQA} task has a zero-shot nature, as is also observed by the authors of Text-KVQA, the 64870 questions in the Text-KVQA dataset have only about 1400 unique answers. In fact, using the 1000 most common answers from a train set comprising $80\%$ split of the data, about $98\%$ of the questions in the test set can be answered. Thus, this improved result can be attributed to the dense answer space indicated by the Vocab UB in Table \ref{tab:dscomp}.

Across models, pre-training on additional tasks and datasets can lead to significant improvements, as is observed in Table \ref{tab:sotapre}.
These datasets, however, comprise additional images and question-answer pairs and require resource-intensive manual annotation. Pre-training also requires additional computing resources. In contrast, our mined external knowledge consists of text descriptions for a subset of scene text words obtained in an unsupervised manner from web scraped data. Further, our mined knowledge facts form part of our feature set for each image and do not require additional training. It is interesting to observe that despite the additional pre-training tasks for TAP (row 4) and additional pre-training datasets for M4C (row 2) and SA-M4C (row 3), our knowledge-enabled model EKTVQA (row 1) still performs at par with them. Finally, we note that the improvements due to TAP (row 6) can be attributed to not just the additional annotated datasets but also improved OCR quality, as is evident by comparing rows 4 and 5.

\vspace{-5pt}
\begin{table}[h]
 \begin{center}
\small
\caption{Upper Bounds and OCR systems:
\textbf{OCR UB}: Percentage of answers comprising OCR tokens alone, \textbf{Vocab UB}: Percentage of answers comprising fixed vocabulary words only, \textbf{OCR + Vocab UB}: Percentage of answers comprising OCR tokens and answer vocabulary words, \emph{i.e.}, the maximum achievable accuracy.}\label{tab:compocr}
 \resizebox{\linewidth}{!}{%
 \begin{tabular}{|l|c|c|c| }\hline
 & \multicolumn{3}{c|}{TextVQA Val accuracy(\%)} \\ \hline
 &Rosetta-ml &Rosetta-en &Google-OCR \\ \hline
 OCR UB &37.12 &44.9 &53.4 \\
 Vocab UB &48.46 &59.02 &57.14 \\
 OCR + Vocab UB &67.56 &79.72 &84.4 \\ \hline
 \end{tabular}}
\end{center} 
\vspace{-5pt}
\end{table}
\vspace{-5pt}
\subsection{\textup{\textbf{How knowledge}} helps}\label{subsec:resqua}
\begin{figure*}
\begin{center}
\includegraphics[width=1\textwidth]{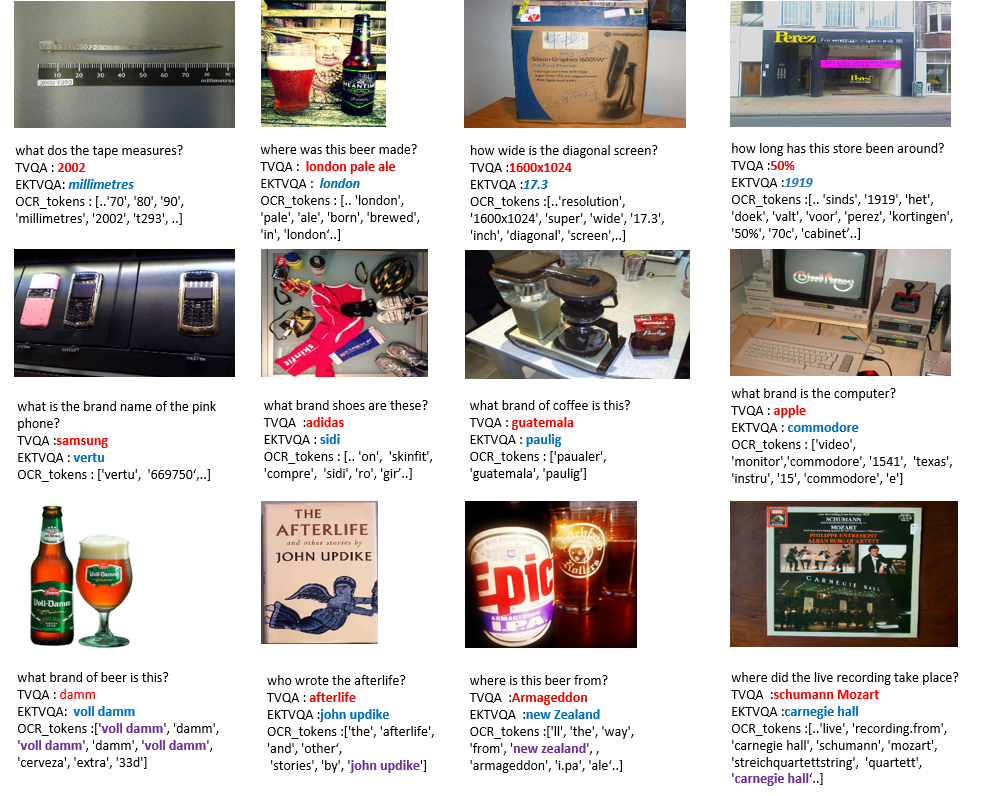}
\caption{ Examples of knowledge effectiveness. TVQA is our baseline similar to M4C~\cite{M4C_CVPR_2020} and does not use external knowledge. The correct answers are marked in Blue, while incorrect ones are marked with Red. Multiword Entities, row 3, are marked with Purple. 
In Row 1, we see observe that with knowledge we are able to improve upon answer entity correctness. Row 2 shows the effectiveness of external knowledge in dealing with Dataset Bias. Finally in Row 3, we showcase our Multi-word Entities.
}\label{fig:knog}
\end{center}
\end{figure*}
\begin{figure*}
\begin{center}
\includegraphics[width=1\textwidth]{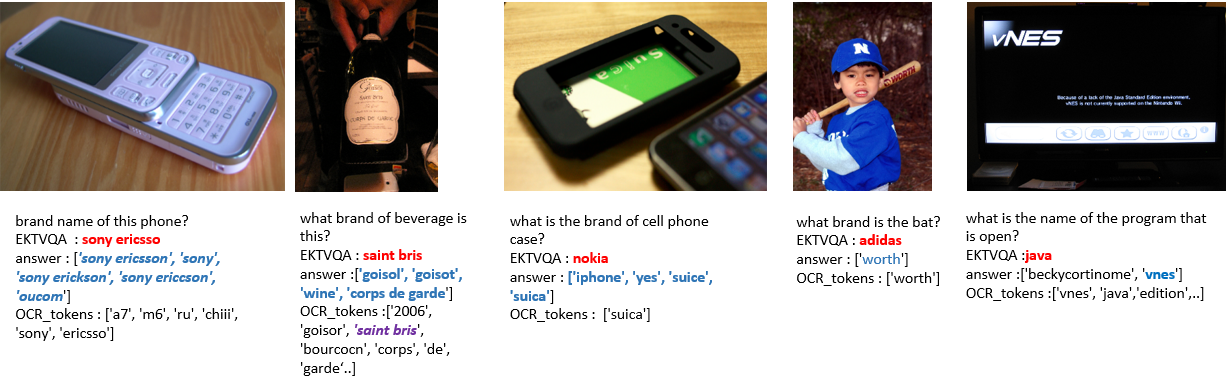}
\caption{ Examples of knowledge limitations. The correct answers are marked in Blue, while incorrect ones are marked with Red. Multi-word Entities are marked with Purple }\label{fig:knob}
\end{center}
\end{figure*}

\subsubsection{ \textup{\textbf{Mitigating dataset bias}}}
Problems due to dataset biases are well studied~\cite{KAFLE20173, Agrawal2017CVQAAC}. This usually results in the network incorrectly predicting the dominant class label for samples from other sparsely populated classes. It can be attributed to data distribution. In \mbox{Text-VQA}, this manifests with the model trying to predict answers it has often encountered in similar image-question contexts, failing to incorporate the OCR tokens that might form the correct answer. 

Row 2 of Figure \ref{fig:knog} demonstrates that despite the presence of the correct answer words in the detected OCR tokens, the baseline TVQA failed to predict the correct answer words and incorrectly predicted semantically similar words from the vocabulary. For example, we can imagine the model has seen the popular brands `samsung' or `apple' more often than say the less known brands `vertu' or `commodore' in the context of phones and computers, respectively, and is thus biased towards predicting them as answers. This problem is more acute due to the zero-shot characteristics of the \mbox{Text-VQA} task, where an unseen OCR token could be an answer word. Our key idea has been that knowledge facts like `Vertu is British Phone Company' or `Commodore is Home Computer Manufacturer' can help us understand and thus reason about lesser-seen or unseen OCR tokens `vertu' or `commodore.' In Row 2 of Figure \ref{fig:knog}, we observe that our EKTVQA model brings awareness about the OCR tokens through the associated external knowledge, which leads to their correct choices as answer words. Thus knowledge helps against bias by bringing in much-needed additional information about lesser-seen or completely unseen OCR tokens.

\subsubsection{\textup{\textbf{Improved Answer Entity type correctness}}} The question asked often pre-determines the type of answer entity.
Entity-type question words like `what time,' `which place,' and `how wide’ point towards distinct answer entity types. In Row 1 of Figure \ref{fig:knog}, we see examples where external knowledge about the OCR tokens enables us to learn this entity type association better. As seen with the responses from TVQA in Row 1, we note that semantic word embedding~\cite{fasttext} and character information~\cite{phoc} about OCR token lead to basic entity type correctness like numeric or non-numeric. However, the knowledge descriptions of the scene text words are typically verbose and detailed, allowing us to better associate the scene text words with related question words.
\vspace{-5pt}
\subsubsection{\textup{\textbf{Multiword Entity Recognition}}} Multiword Entities are entities with multiple words in the name, e.g. `New York,' `Bob Dylan,' `Into The Wild,' etc. We detect multiword entities during the external knowledge extraction phase, where we bind together OCR tokens if they are part of the same-named entity. E.g., if for OCR token `York,' we find a returned named entity `New York' from GKB, and the OCR token `New' is in the neighborhood, we bind them as `New York.'

In Row 3 of Figure \ref{fig:knog}, we see examples of correct answers using multiword entities. For popular words like `New York,' our trained multimodal transformer itself should be enough to predict them in sequence, but for a lesser-seen couplet of words such as `voll damm,' such binding is more beneficial. 

\vspace{-5pt}
\subsection{\textup{\textbf{Ablation Studies}}}\label{subsec:resabl}
In this section, we report results on ablated instances of our model across datasets to highlight the contribution of individual model components and draw insights about the transferable properties of external knowledge. 

\begin{table}[h]
\setcounter{magicrownumbers}{0} 
 \begin{center}
\small
\caption{Ablation study 1: Role of Knowledge related Components}\label{tab:ablcomp}
 \resizebox{\linewidth}{!}{
 \begin{tabular}{|l|l|c|c|c|c|}\hline
 \multicolumn{4}{|c|}{} &\multicolumn{1}{|c|}{TextVQA } & \multicolumn{1}{|c|}{ST-VQA} \\ \hline
 & Model & Validation & Reasoning &Val. Acc. &Val. ANLS. \\ \hline
\rownumber & TVQA & - & - & 41.7 & 0.4893\\
\rownumber & EKTVQA$_{UnC}$ & Contextual & Unconstrained & 42.73 & 0.4875 \\
\rownumber & EKTVQA$_{Rnd}$ & Random & Constrained & 42.89 & 0.4893 \\ 
\rownumber & EKTVQA$_{All}$ & Sum of All & Constrained & 43.54 & 0.4856 \\ 
\rownumber & EKTVQA & Contextual & Constrained & 44.26 & 0.4991 \\ \hline
\end{tabular}}
\end{center} 
\end{table}
\vspace{-5pt}
\subsubsection{\textup{\textbf{Ablation study 1: \textbf{\small Role of Model Components}}}} 
We propose validating candidate knowledge facts through image context, and reasoning with this validated knowledge through constrained interaction. In this section, we investigate the individual role played by these two knowledge components by comparing our full model with ablated instances. We briefly describe the ablated models below:

\paragraph{\bfseries\scshape EKTVQA} EKTVQA is our full model, which uses context to validate and thus select knowledge, and enforces constrained interaction during reasoning with the knowledge nodes. 
\vspace{-10pt}
\paragraph{\bfseries\scshape TVQA} As described in Section~\ref{subsec:ressota} TVQA is our non-knowledge baseline similar to M4C. It does not use any knowledge nodes and thus is devoid of all of our knowledge pipeline components.
\vspace{-10pt}
\paragraph{\bfseries\scshape EKTVQA$_{UnC}$} We propose the use of constrained interaction, as detailed in Section~\ref{subsec:modeladj} to preserve knowledge legitimacy. In EKTVQA$_{UnC}$, we allow unconstrained interaction between the knowledge and text nodes through the use of a zero matrix as the attention Mask \textbf{A}. The remaining components are the same as the EKTVQA full model, including the context-based validation.
\vspace{-10pt}
\paragraph{\bfseries\scshape EKTVQA$_{Rnd}$} In Section~\ref{para:knwextvld} we argue for the use of context to validate and select from amongst candidate meanings. EKTVQA$_{Rnd}$ is our ablated instance, which does not use context but instead randomly selects from among the candidate meanings. It is similar to EKTVQA in every other respect, including the constrained interaction.
\vspace{-10pt}
\paragraph{\bfseries\scshape EKTVQA$_{All}$} This is our ablated instance which, unlike EKTVQA or EKTVQA$_{Rnd}$ does not select one particular candidate meaning, instead aggregates all candidate meanings. Our implementation achieves this by taking a sum of the BERT embeddings of all candidates instead of selecting one.

\vspace{5pt}
As shown in Table \ref{tab:ablcomp}, for both datasets, the inclusion of knowledge leads to improvement over the M4C-based non-knowledge baseline TVQA. We see this improvement to be more pronounced for the \mbox{TextVQA} dataset compared to ST-VQA because the former has more knowledge entries, as is evident in Table \ref{tab:dscomp}. 
Further, we demonstrate the effectiveness of our constrained interaction through its improvements upon the unconstrained baseline EKTVQA$_{UnC}$. As discussed in Section~\ref{subsec:modeladj}, our constrained interaction enforces knowledge legitimacy, enabling it to discriminate amongst OCR tokens based on their corresponding knowledge node. 
Next, we highlight the improvements through context-based candidate knowledge fact validation and selection by comparing with two baselines, EKTVQA$_{Rnd}$, which selects a retrieved candidate knowledge fact randomly, and EKTVQA$_{All}$, which makes no selection and instead settles for aggregation. The improvements shown validate our hypothesis that context can be effectively utilized to select valid candidates that help downstream reasoning.

\subsubsection{\textup{\textbf{Ablation study 2: \textbf{\small Transferable property of external knowledge}}}}
We propose using externally mined knowledge to augment our understanding of scene text nodes. In this section, we investigate the transferable property of the external knowledge by comparing our full model with the ablated instances below:
\vspace{-5pt}
\paragraph{\bfseries\scshape TVQA$^{*}$} TVQA$^{*}$ is the TVQA model loaded with EKTVQA trained weights, mimicking a model that has seen external knowledge during training, but does not use knowledge during testing.
\vspace{-5pt}
\paragraph{\bfseries\scshape EKTVQA$^{*}$} EKTVQA$^{*}$ is the EKTVQA model loaded with TVQA trained weights. It mimics a model that is not trained with knowledge but can use knowledge during testing. 

\vspace{5pt}
In Table \ref{tab:ablcomp_knw} rows 1-2, we observe that even if the use of external knowledge is limited to the training phase, we obtain significant improvements in the test results. This is in line with our previous hypothesis in Section~\ref{subsec:resqua}, about how the verbose knowledge descriptions lead to better generalization. Thus, while our model uses no explicit instance-specific knowledge during testing, the learned knowledge-enabled semantics from the training phase lead to the improvement. Rows 3 and 4 suggest that knowledge cannot simply be plugged in only at test time, and learning the text-knowledge association, enabled through our knowledge components, is essential.

\begin{table}[h]
\setcounter{magicrownumbers}{0} 
 \begin{center}
\small
\caption{Ablation study 2: Transferable property of external knowledge }\label{tab:ablcomp_knw}
 \resizebox{\linewidth}{!}{
 \begin{tabular}{|l|l|c|c|c|c|c}\hline
 & &\multicolumn{2}{|c|}{Knowledge} &\multicolumn{1}{|c|}{TextVQA} & \multicolumn{1}{|c|}{ST-VQA } \\ \hline
 & Model &Train & Test &Val. Acc. &Val. ANLS. \\ \hline
\rownumber & TVQA & $\times$ & $\times$ & 41.7 & 0.4893 \\
\rownumber & TVQA$^{*}$ &$\checkmark$ & $\times$ & 43.12 & 0.4908 \\
\rownumber & EKTVQA$^{*}$ &$\times$ & $\checkmark$ & 39.36 & 0.4635 \\ 
\rownumber & EKTVQA &$\checkmark$ & $\checkmark$ & 44.26 & 0.4991 \\ \hline 
\end{tabular}}
\end{center} 
\end{table}

\subsection{Bottleneck and Limitations}

\subsubsection{\textup{\textbf{Knowledge effectiveness depends on OCR quality}}}
Valid knowledge extraction is dependent on correct OCR recognition. Incorrect OCR recognition can lead to either a complete miss or false positives in the knowledge extraction phase. In the first and second images of Figure \ref{fig:knob}, the incorrect OCR tokens `goisol' (`goisot') and `ericsso' (`ericsson') cannot be helped by external knowledge, which subsequently leads to incorrect answer predictions in both cases. As discussed in Section~\ref{subsec:datasets}, we observe that GKB has tolerance towards incorrect spellings, wherein it can return correct meanings even with incorrect OCR tokens. So while in the case of `goisot,' we received no hits, for `ericsso,' GKB tolerance allowed us to get the meaning corresponding to `ericsson.'
However, this does not help as the final answer words are selected from the OCR tokens, as stated in Section~\ref{subsec:ansgen}. Thus despite having the correct meaning corresponding to `ericsson,' our predicted answer word was the incorrect OCR token `ericsso.'

\subsubsection{\textup{\textbf{Knowledge is not complete}}}
Incompleteness in knowledge can manifest itself in two ways, either through a \emph{complete miss} or through \emph{false positives} by incorrectly referring to another entity bearing the same name as the queried text. Thus knowledge incompleteness is a significant cause of failure, as illustrated in Figure \ref{fig:knob}. For the fourth and fifth images, the correct answer forming OCR tokens `worth' and `vnes' were absent in the external knowledge base. While for the third image, the OCR token `suica' refers to a smart card or a village in Bosnia, according to GKB, failing to include any reference to a cell phone case.

\subsection{\textup{\textbf{Results Summary}}}
Our main takeaway is that external knowledge leads to improved generalization and allows us to deal with dataset bias. Our primary results in Table \ref{tab:sotacomp} and \ref{tab:sotaEKTVQA} show that, given the same upstream OCR system, and training data, our results match the state-of-the-art on three public datasets. Furthermore, Table \ref{tab:sotapre} suggests our model~(Row 1) is at par with models trained on additional datasets~(Rows 2,3) and even those incorporating pre-training tasks~(Rows 4,5).

Through our experiments, we also justify our model components. We identify contextual validity check as an important subtask for noisy web scraped knowledge facts and justify it through our results (Table \ref{tab:ablcomp} Rows 3,4 against Row 5). Our arguments for node-to-node knowledge legitimacy achieved through constrained interaction are validated by our results in Table \ref{tab:ablcomp} (Rows 2 against Row 5). Finally, our results also identify OCR quality and knowledge incompleteness as major bottlenecks to overcome in the future. 

Apart from Question Answering, multimodal transformers have also been applied to a range of other downstream tasks like Captioning~\cite{sidorov2020textcaps}, Question Generation, Vision and Language Navigation (VLN)~\cite{chen2021history} which requires an understanding of scene text and can thus benefit from the additional knowledge facts. Thus our knowledge-enabled multimodal transformer module can be readily adapted to these existing tasks leading to further improved results without additional training data.

\section{Conclusion}\label{sec:concl}
In this work we look at the use of mined external knowledge applied to the open-ended \mbox{Text-VQA} task. Through our experiments we observe that this external knowledge not only provides invaluable information about unseen scene text elements but also augments the understanding of the text in general with detailed verbose descriptions. Our knowledge-enabled model is robust to novel text, predicts answers with improved entity type correctness, and can even recognize multiword entities. However, the knowledge pipeline is susceptible to erroneous OCR tokens, which can lead to false positives or complete misses. This also explains how our performance on the datasets is correlated with the particular OCR systems used. 

While the use of knowledge in Vision Language understanding tasks has been proposed in the past, we propose an end-to-end pipeline to extract, validate and reason with noisy web scraped external knowledge. We present this as a complementary approach to training on additional annotated data or pre-training tasks to achieve generalization and deal with dataset bias. Finally, we hope our results inspire the community to incorporate external knowledge pipelines into their works and gain improvements in results.

{\small
\bibliography{egbib2}
}
\begin{IEEEbiography}[{\includegraphics[width=1in,height=1.25in,clip,keepaspectratio]{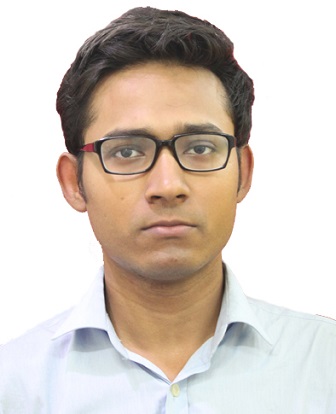}}]{Arka Ujjal Dey} received his Bachelor's and Master's Degrees in Computer Science from Calcutta University in 2010, and 2012 respectively. In 2015 he obtained his Master of Technology Degree from IIT Jodhpur, where he is currently enrolled as a Ph.D. student in the Computer Science and Engineering Department. His research interests lie at the intersection of Computer Vision and Natural Language Processing. 
\end{IEEEbiography}
\vspace{-200pt}
\begin{IEEEbiography}[{\includegraphics[width=1in,height=1.25in,clip,keepaspectratio]{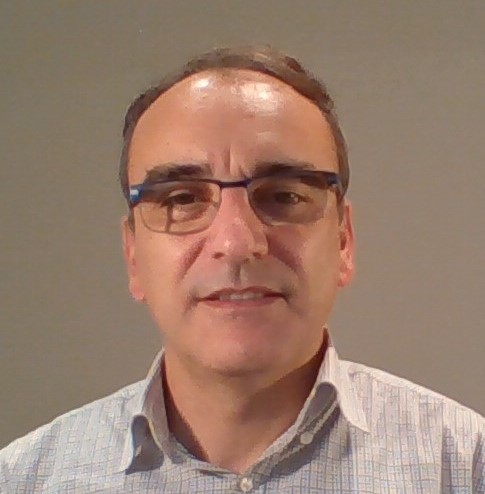}}]{Ernest Valveny} received the Ph.D. degree in 1999 from the Universitat Autònoma de Barcelona (UAB), where he is currently an associate professor and member of the Computer Vision Center (CVC). His research work focuses mainly on document analysis and pattern recognition, and more specifically, in the fields of robust reading, text recognition and retrieval, and document understanding. He has published more than 20 papers in international journals and more than 100 papers in international conferences. He is a member of IAPR and of the editorial board of the International Journal on Document Analysis and Recognition and has served as a reviewer for many international journals and conferences.
\end{IEEEbiography}
\vspace{-200pt}
\begin{IEEEbiography}[{\includegraphics[width=1in,height=1.25in,clip,keepaspectratio]{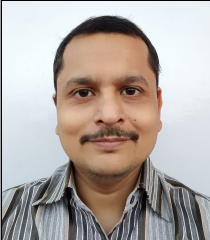}}]{Gaurav Harit} received a Bachelor of Engineering degree, in Electrical Engineering, from M.B.M. Engineering College, Jai Narayan Vyas University Jodhpur, in 1999. He received a Master of Technology and Ph.D. degrees from IIT Delhi, in 2001 and 2007, respectively. He has been a faculty member in the Department of Computer Science and Engineering at the Indian Institute of Technology Jodhpur since 2010. Before joining IIT Jodhpur, he was a faculty member in the Department of Computer Science and Engineering at IIT Kharagpur. His areas of interest are Document Image Analysis and Video Analysis.
\end{IEEEbiography}

\EOD

\end{document}